\documentclass{article}
\usepackage[final]{neurips_2024}

\usepackage[utf8]{inputenc} % allow utf-8 input
\usepackage[T1]{fontenc}    % use 8-bit T1 fonts
\usepackage{hyperref}       % hyperlinks
\usepackage{url}            % simple URL typesetting
\usepackage{booktabs}       % professional-quality tables
\usepackage{amsfonts}       % blackboard math symbols
\usepackage{nicefrac}       % compact symbols for 1/2, etc.
\usepackage{microtype}      % microtypography
\usepackage{xcolor}         % colors
\usepackage{graphicx}

\usepackage{amsmath}
\usepackage{amssymb}
\usepackage{bm}
\usepackage{graphicx}
\usepackage{siunitx}
\usepackage{natbib}
\usepackage{wrapfig}
\usepackage[percent]{overpic}
\usepackage{amsthm}

\usepackage{float}

\title{What exactly has TabPFN learned to do?}
\author{
Calvin McCarter \\
\texttt{mccarter.calvin@gmail.com}
}
\date{May 2024}

\begin{document}

\maketitle

\begin{abstract}
TabPFN [Hollmann et al., 2023], a Transformer model pretrained to perform in-context learning on fresh tabular classification problems, was presented at the last ICLR conference. To better understand its behavior, we treat it as a black-box function approximator generator and observe its generated function approximations on a varied selection of training datasets. Exploring its learned inductive biases in this manner, we observe behavior that is at turns either brilliant or baffling. We conclude this post with thoughts on how these results might inform the development, evaluation, and application of prior-data fitted networks (PFNs) in the future.\footnote{The initial version of this manuscript was published in the ICLR 2024 Blogposts Track, and is available at \url{https://iclr-blogposts.github.io/2024/blog/what-exactly-has-tabpfn-learned-to-do/}. This version includes, in its appendix, the results of a 2025 re-analysis on TabPFN V2 \cite{hollmann2025accurate}. Code is available at \url{https://github.com/calvinmccarter/tabpfn-eval}.}
\end{abstract}

\section{Introduction}

TabPFN \citep{hollmann2023tabpfn} is a deep learning model pretrained to perform in-context learning for tabular classification. 
Since then, it has attracted attention both for its high predictive performance on small dataset benchmarks and for its unique meta-learning approach.
This meta-learning approach, which builds upon earlier work on prior-data fitted networks (PFN) \citep{muller2022transformers}, requires only synthetically-generating data: structural causal models (SCMs) \citep{pearl2009causality} are randomly generated, then training datasets are sampled from each SCM.
On fresh classification tasks, no training (i.e. weight updating) is needed; instead, training data is given as context to TabPFN, a Transformer \citep{vaswani2017attention} model with self-attention among training samples and cross-attention from test samples to training samples.
TabPFN can be optionally used with ensembling, wherein the forward pass is repeated with random permutations of features and class labels, and with power transformation applied to random subsets of features.
Subsequent works have reproduced its classification performance on other tabular benchmarks \citep{mcelfresh2023neural}, and analyzed its theoretical foundations \citep{nagler2023statistical}.

At the same time, TabPFN has received criticism from within the applied ML community, around concerns that its ``one large neural network is all you need'' approach is fundamentally flawed and that its performance on public benchmarks may be due to overfitting \citep{tunguz,manokhin}.

In this article, we will attempt to demystify TabPFN's behavior in order to move towards a resolution to these questions.
With this goal, we will take a different tack to analyzing TabPFN than previous works: 
we will neither theoretically analyze its meta-learning pre-training approach, nor run it on yet another dataset-of-datasets, nor even mechanistically interpret the meaning of specific model weights or subnetworks. 

Instead, we will first explore its holistic behavior on two simple settings, in order to develop an intuition about TabPFN as a function approximation generator.
This is motivated by the observation that TabPFN once fitted on fresh training data (even though ``fitting'' is merely storing the training data), is not mathematically different from any other fitted model: it is simply a function $f_{\mathcal{D}, \theta}: x \rightarrow y$ from test input $x$ to prediction $y$,
where $\mathcal{D} = (X_{\textrm{train}}, y_{\textrm{train}})$ is the training data and $\theta$ are the TabPFN model weights.
By plotting $f$ for various case studies of $(X_{\textrm{train}}, y_{\textrm{train}})$, we aim to better understand what statistical knowledge has been represented in model parameters $\theta$.

Next, we will evaluate TabPFN on two non-standard tabular ML classification tasks, comparing its performance with other methods.
These atypical tasks can be thought of as out-of-distribution relative to the synthetic pretraining datasets upon which TabPFN was pretrained.
This analysis will help indicate whether TabPFN was overfit to the statistical peculiarities of publicly-available small tabular datasets, or whether it has learned generalizable principles that lead to sensible behavior even in out-of-domain settings.

\section{1d binary classification}

We begin by examining the case of binary classification with 1d inputs. To better illustrate the inductive biases of the base TabPFN model, we do not use ensembling in this section unless otherwise indicated. 

Below in Figure \ref{fig:plusminus1-nonmonotone}, we show the predictions for two training samples located at $\textcolor{green}{+1}$ and $\textcolor{red}{-1}$, labeled green and red, respectively. We see that the probabilities are non-monotonic, as one would obtain from a sigmoid function; not only do we see that the model has higher uncertainty on the far sides of the training points, we see that between them there is a small wiggle. We also see that the decision boundary biased below 0.5; likely this is because TabPFN has learned that features are have right-skewed distributions.

\begin{figure}[H]
    \centering
    \includegraphics[scale=0.8]{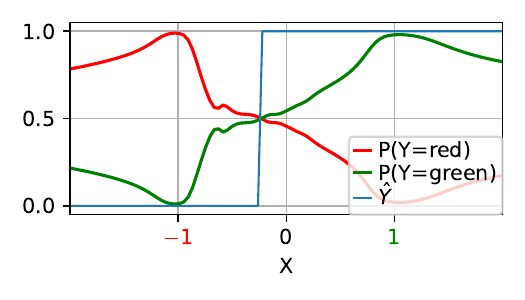}
    \vspace{-0.2in}
    \caption{TabPFN predicted probabilities for simple 1d scenario, with data in red and green.}
    \label{fig:plusminus1-nonmonotone}
\end{figure}

These wiggles and asymmetry more-or-less disappear once we incorporate ensembling, shown below in Figure \ref{fig:plusminus1-ensembles2}.
However, the general shape of the predicted probability function is similar regardless of the number of ensembles.

\begin{figure}[H]
    \centering
    \includegraphics[scale=0.8]{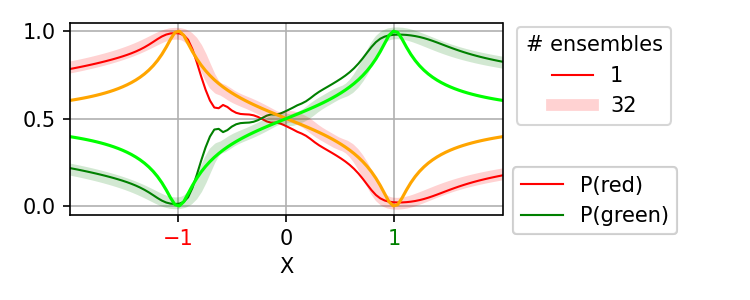}
    \vspace{-0.2in}
    \caption{TabPFN predicted probabilities for simple 1d scenario, for varying number of ensembles. Also shown are the predicted probabilities from using inverse-square-root of Euclidean distance within softmax, in orange and lime-green.}
    \label{fig:plusminus1-ensembles2}
\end{figure}
The above results raise the question of what parametric attention function might have been learned by TabPFN.
No simple dot-product-based or Euclidean distance-based function (used within the softmax operation) exactly recapitulated the observed predicted probabilities.
However, the general shape of inverse-square-root of Euclidean distance matched reasonably well, particularly between the two training points.
Still, it appears that TabPFN has meta-learned an attention function that outperforms previously-known attention functions on small datasets.

Next, we look at the effect of duplicating features. We tried repeating the +1 and -1 inputs for a total of 1, 4, 16, and 64 copies, as shown below in Figure \ref{fig:plusminus1-repeat-features}. The effect is to push the predicted probabilities away from 0.5, although we observe  diminishing marginal effects as the number of repeats increases.

\begin{figure}[H]
    \centering
    \includegraphics[scale=0.8]{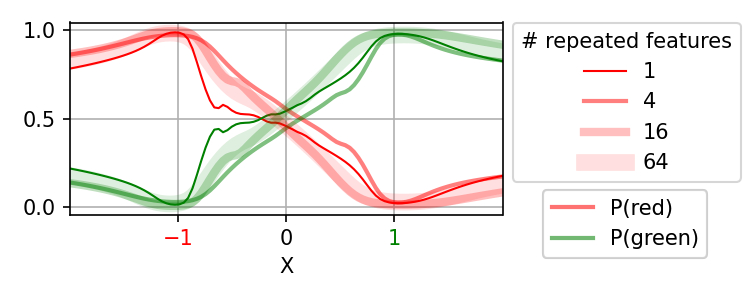}
    \vspace{-0.2in}
    \caption{TabPFN predicted probabilities for simple 1d scenario, but with repeated features.}
    \label{fig:plusminus1-repeat-features}
\end{figure}

\begin{figure}[H]
    \centering
    \includegraphics[scale=0.8]{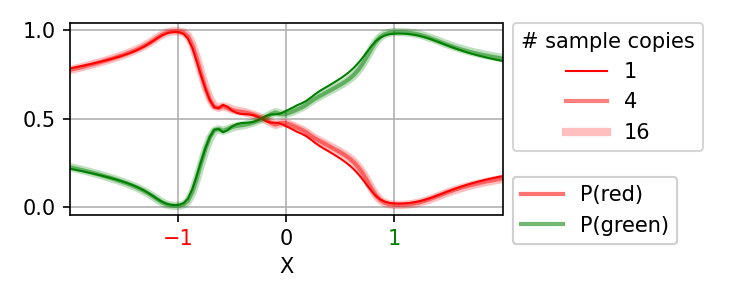}
    \vspace{-0.2in}
    \caption{TabPFN predicted probabilities for simple 1d scenario, but when both red and green samples are duplicated.}
    \label{fig:plusminus1-repeat-samples}
\end{figure}

Meanwhile, there is no discernible effect from replicating samples, when both red and green samples are replicated. Below, in Figure \ref{fig:plusminus1-repeat-samples}, we show the predicted probabilities, when both red and green samples are each copied for a total of 1, 4, 16, and 64 times.

\begin{figure}[H]
    \centering
    \includegraphics[scale=0.8]{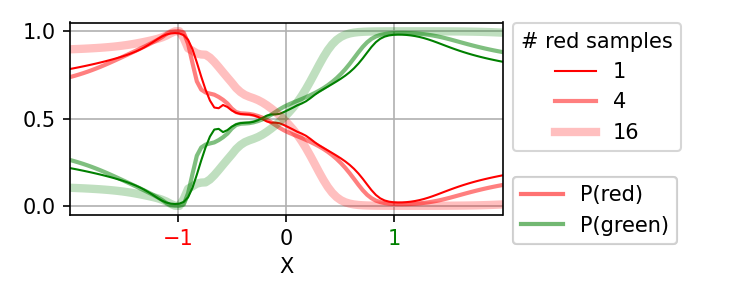}
    \vspace{-0.2in}
    \caption{TabPFN predicted probabilities for simple 1d scenario, but when the red sample is duplicated.}
    \label{fig:plusminus1-repeat-red}
\end{figure}

In contrast, there is an impact to repeating only the red sample.
Below, in Figure \ref{fig:plusminus1-repeat-red} is shown the effect of repeating only the red sample.
While this unsurprisingly increases the probability of red for $X < 0$, it bizarrely increases the probability of green for $X > 0$.
This is especially strange because repeating green samples in the previous setting did not have the same effect.
This behavior of TabPFN seems suboptimal; it remains to be seen whether this behavior was optimal for its pretraining data, or whether this is some kind of artifact of TabPFN's architecture or training.

Finally, we were unable to find evidence that TabPFN is able to detect periodic patterns in the training data, as exemplified for three different training patterns shown below in Figure \ref{fig:plusminus1-periodic}. 
This behavior of TabPFN suggests that it does not support either periodic interpolation or extrapolation.
Furthermore, we observe that as the number of observed cycles in the data increases, the predicted probabilities trend toward 0.5, which also seems suboptimal.
We also notice that there is marked left-right asymmetry in these settings.

\begin{figure}[H]
    \centering
    \includegraphics[scale=0.7]{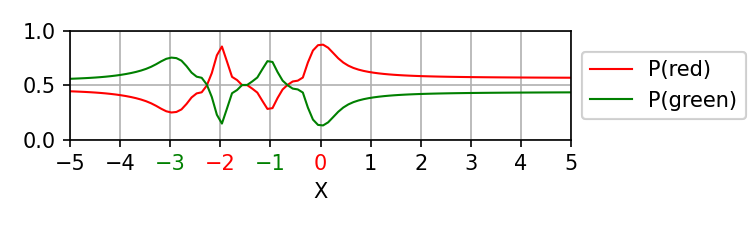} \\
    \includegraphics[scale=0.7]{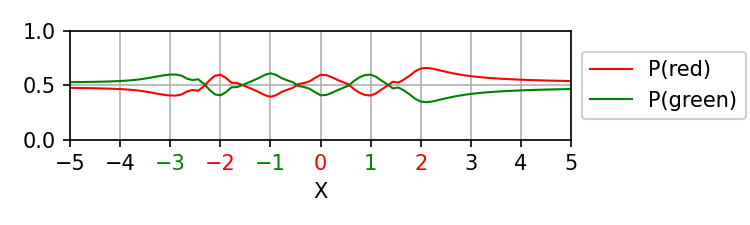} \\
    \includegraphics[scale=0.7]{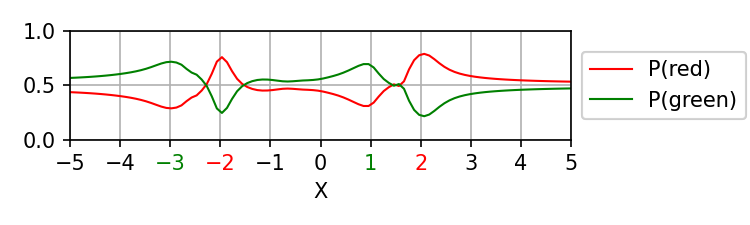}
    \vspace{-0.2in}
    \caption{TabPFN predicted probabilities for three scenarios with periodic patterns.}
    \label{fig:plusminus1-periodic}
\end{figure}

\section{2d multiclass classification}

Here, we examine the behavior of TabPFN on 2d input data, on problems with as many samples as classes.
Below, in Figure \ref{fig:voronoi}, we show results for both randomly-spaced and grid-spaced inputs, and for both ensembling and no-ensembling settings of TabPFN.
In each plot, we show the training data, their corresponding Voronoi diagrams, and finally the model predictions for the test inputs.
We see that, without ensembling, TabPFN performs quite poorly, partitioning the input space in a non-sensical manner.
The results markedly improve when we use 32 ensembles.
Particularly for the randomly-spaced training points, the model predictions clearly resemble the Voronoi diagram, suggesting that (ensembled) TabPFN has meta-learned to perform 1-nearest-neighbor classification in the setting where each class has a single training sample.

On the other hand, that this behavior relies upon ensembling suggests that the base TabPFN model could be further improved.
In the original paper, \cite{hollmann2023tabpfn} express the hope that a future better version of TabPFN would not need to rely upon ensembling for permutation invariance, by having internalized that behavior through better architecture and training.
The aforementioned observed behavior suggests that ensembling improves performance not only by (approximately) enforcing permutation invariance, but also by producing lower variance estimators; if so, the base model could also be trained to do the latter directly.

\begin{figure}[H]
    \centering
    \includegraphics[scale=0.5]{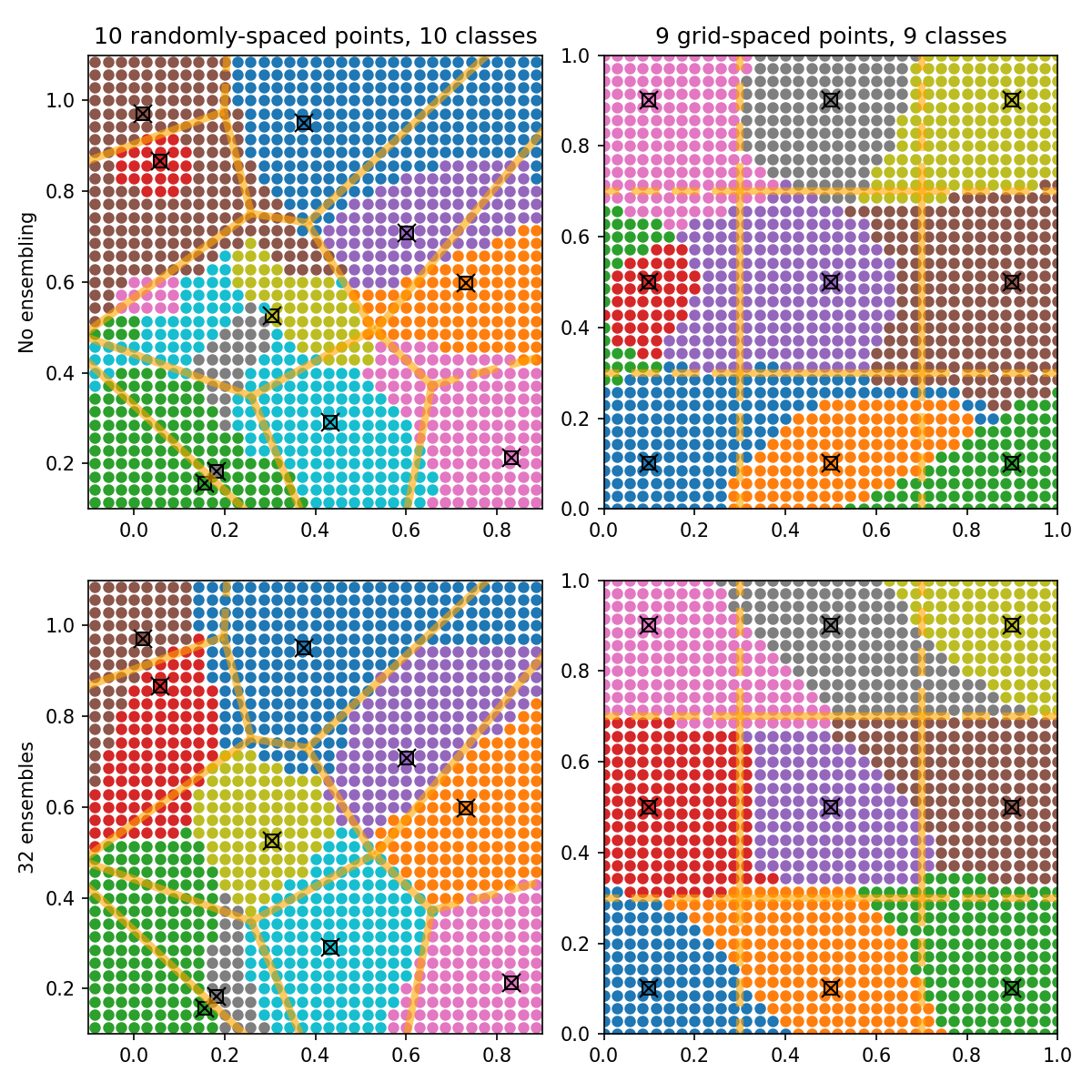}
    \vspace{-0.2in}
    \caption{TabPFN predictions on randomly-spaced points (left) and grid-spaced points (right). The training points are depicted as $\times$s. The yellow lines depict the Voronoi diagram of the training points. The test points are colored by TabPFN's predictions, using the same color scheme as the training points. We see that, without ensembling, TabPFN's predicted classes do not form contiguous regions over the input space.}
    \label{fig:voronoi}
\end{figure}

\iffalse
\setlength{\tabcolsep}{1pt}
\begin{figure}[H]
    %\centering
   (A) Without ensembling \vspace{-20pt} \\
   \begin{center}
   \begin{tabular}{cc}
        %10 randomly-spaced points, 10 classes & 9 grid-spaced points, 9 classes \\
        \includegraphics[scale=0.5]{figures/voronoi.pdf} & \includegraphics[scale=0.5]{figures/voronoi-grid.pdf}
   \end{tabular} \\       
   \end{center}
    (B) With ensembling \vspace{-20pt} \\
    \begin{center}
    \begin{tabular}{cc}       
        \includegraphics[scale=0.5]{figures/voronoi-32.pdf} & \includegraphics[scale=0.5]{figures/voronoi-grid-w32.pdf}
    \end{tabular} \\
    \end{center}
    \vspace{-0.2in}
    \caption{TabPFN predictions on 10 randomly-spaced points with 10 classes (left), and 9 grid-spaced points with 9 classes (right). The training points are depicted as $\times$s. The yellow lines depict the Voronoi diagram of the training points. The test points are colored by TabPFN's predictions, using the same color scheme as the training points. We see that, without ensembling, TabPFN's predicted classes do not form contiguous regions over the input space.}
    \label{fig:voronoi}
\end{figure}
\fi 

\section{Cancer status classification from high-dimensional gene expression}

We now turn towards a comparison of TabPFN with logistic regression (LR), support vector classification (SVC), and XGBoost \citep{chen2016xgboost} on the \textit{BladderBatch} \citep{leek2016bladderbatch} cancer status classification task.
The \textit{BladderBatch} dataset consists of 57 samples, 22{,}283 gene expression features, and 3 classes (``normal'' vs ``biopsy'' vs ``cancer''). 
This is an extremely high-dimensional problem compared to TabPFN's intended use for $d \le 100$; also, linear models tend to be sufficient for predicting cancer status given gene expressions.
Thus, this setting is far outside the domain on which we would expect TabPFN to perform well, particularly if it had been overfit to small tabular datasets.
Furthermore, the 57 samples come from 5 different batches of gene microarray measurements. 
This adds additional difficulty to the task, because there is confounded shift between the technical batch effect and the unequal proportions of cancer status in the different batches \citep{mccarter2022towards}.

For all methods, we do not perform hyperparameter search, in order to simulate the scenario where there are too few samples to perform cross-validation without the risk of overfitting.
We use the scikit-learn \citep{pedregosa2011scikit} implementations of LR and SVC with their default hyperparameters.
For TabPFN, we use the default hyperparameter of 32 ensembles; we also enable feature subsampling as is required for $d > 100$ problems.

Results are shown below in Figure \ref{fig:bladderbatch-comparison}, aggregated over 10 random 75-25 train-test splits, and evaluated via both accuracy and macro-averaged F1-score.
TabPFN has a surprisingly strong showing, handily beating SVC and XGBoost, while almost matching logistic regression.
This pattern holds both when we use all features and also when we use only the first 1k features.
\begin{figure}[H]
    \centering    
    \includegraphics[scale=0.17]{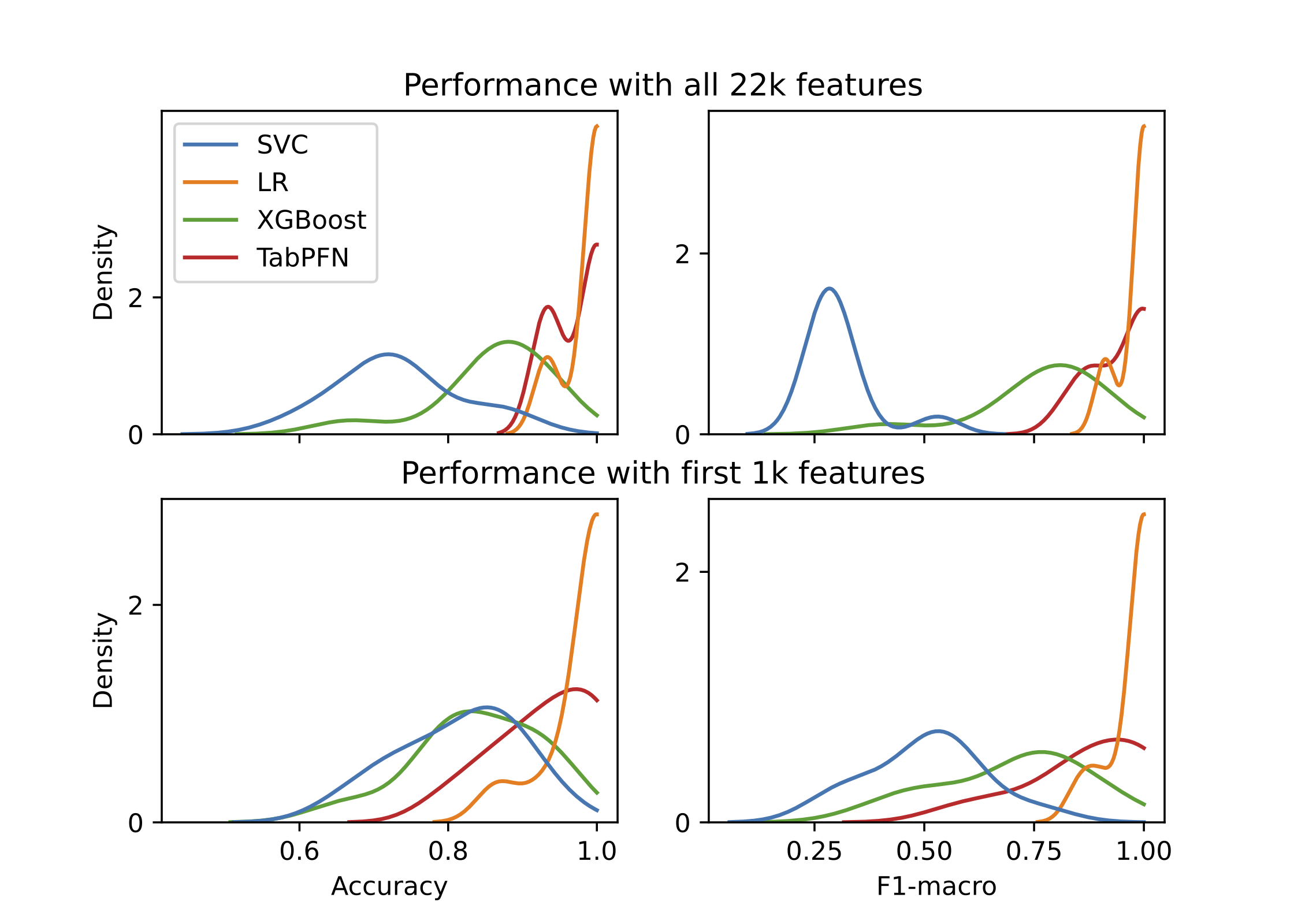}
    \caption{Results on \textit{BladderBatch} cancer classification.}
    \label{fig:bladderbatch-comparison}
\end{figure}

We also evaluate the different methods on a more realistic setting, where we train on 4 out of 5 batches of data and evaluate on all samples from the remaining unseen batch.
Results are shown below in Figure \ref{fig:bladderbatch-batch}. 
While all methods perform worse in this setting, TabPFN still almost matches LR while beating the other baselines.
\begin{figure}[H]
    \centering    
    \includegraphics[scale=0.7]{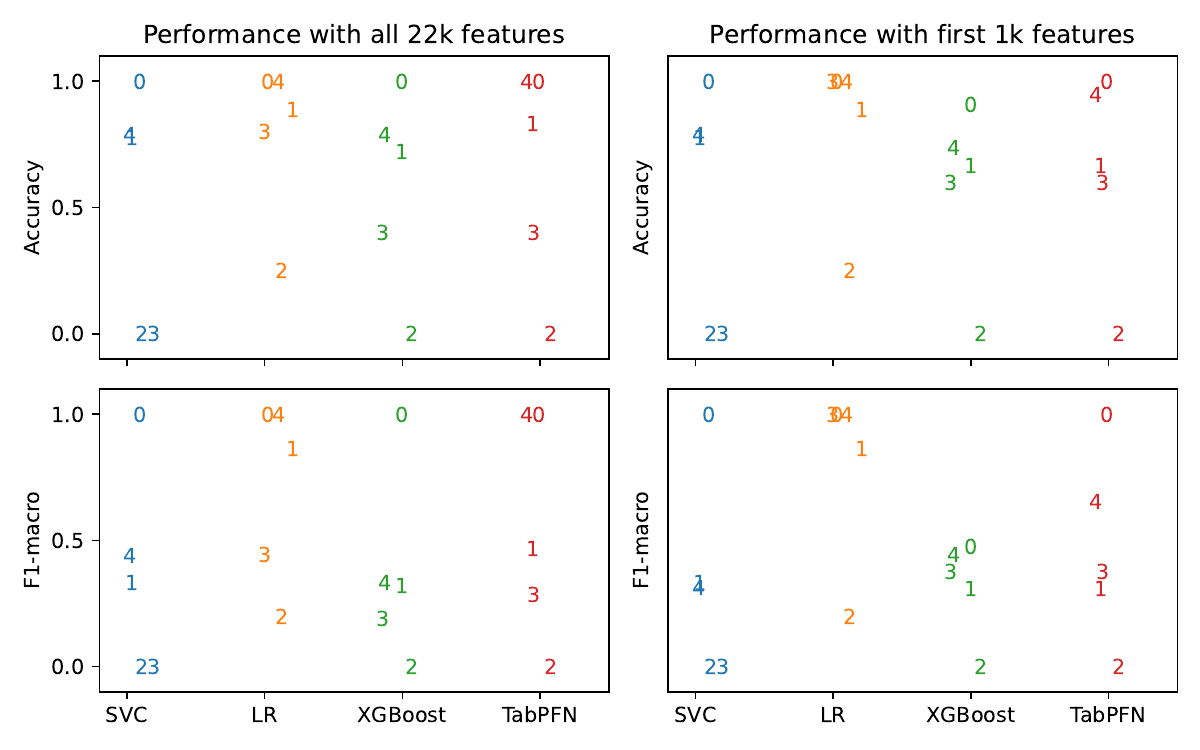}
    \caption{Results on \textit{BladderBatch} cancer classification, for generalizing to other batches. Scatterplot labels are used to indicate the identity of the test batch.}
    \label{fig:bladderbatch-batch}
\end{figure}

We also verify that TabPFN is not simply memorizing the class imbalance in favor of cancer.
We compute confusion matrices, shown below in Figure \ref{fig:bladderbatch-cm}, for each train-test split.
Even though cancer is the most common class in every training split, there does not appear to be any systematic bias across the splits in favor of predicting cancer.
\begin{figure}[H]
    \centering    
    \includegraphics[scale=0.6]{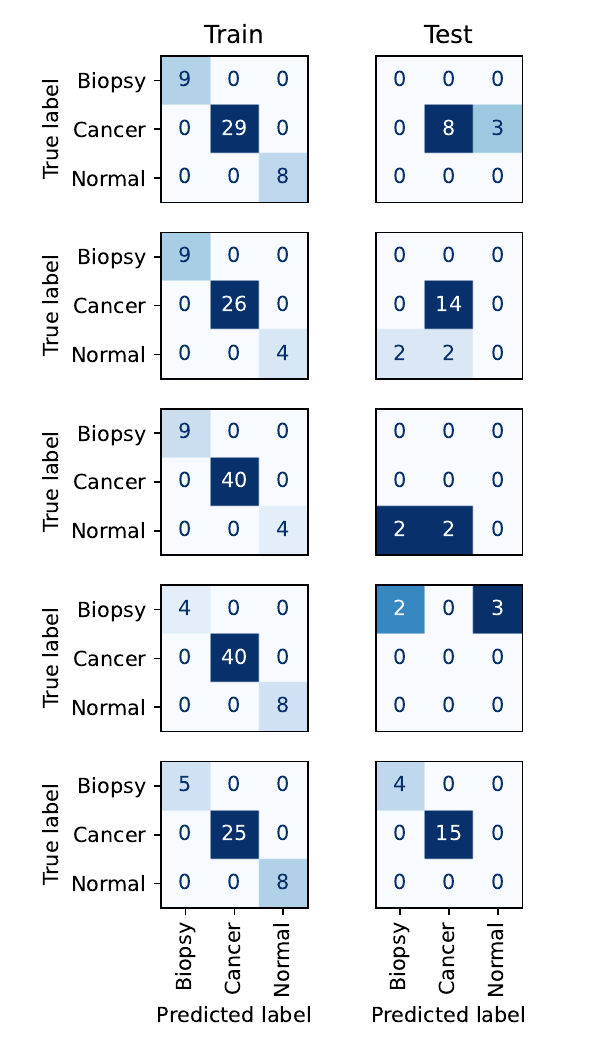}
    \caption{Confusion matrices for \textit{BladderBatch} experiment.}
    \label{fig:bladderbatch-cm}
\end{figure}

\section{Computer vision as a tabular classification problem}

Finally, we compare TabPFN with other methods on two computer vision (CV) tasks.
As in the previous section, we use the default hyperparameter settings for all methods.
We treat MNIST and CIFAR-10 as tabular ML problems with $28*28^2$ and $3*32^2$ features, respectively.
We aggregate over 10 train-test splits, where the test set is the full MNIST / CIFAR-10 test set, and the training set is a random subsample of size 30, 100, 300, and 1000.
Results are shown below in Figure \ref{fig:cv}.
In this experiment, TabPFN was competitive for smaller training set sizes, but lagged as we trained on more samples.
Interestingly, while for cancer classification SVC performed poorly, it performed well for large sample sizes on the CV tasks. 
Meanwhile, while logistic regression (LR) performed well on cancer classification, it struggled in the current setting.
It remains to be seen whether the shared behavioral characteristics of TabPFN and LR in these tasks hold more generally. 
If so, this could motivate future work on meta-learning TabPFN to perform robust classification with a hinge-type loss.

\begin{figure}[H]
    \centering   
    \begin{tabular}{cc}
        MNIST & CIFAR-10 \\
         \includegraphics[scale=0.5]{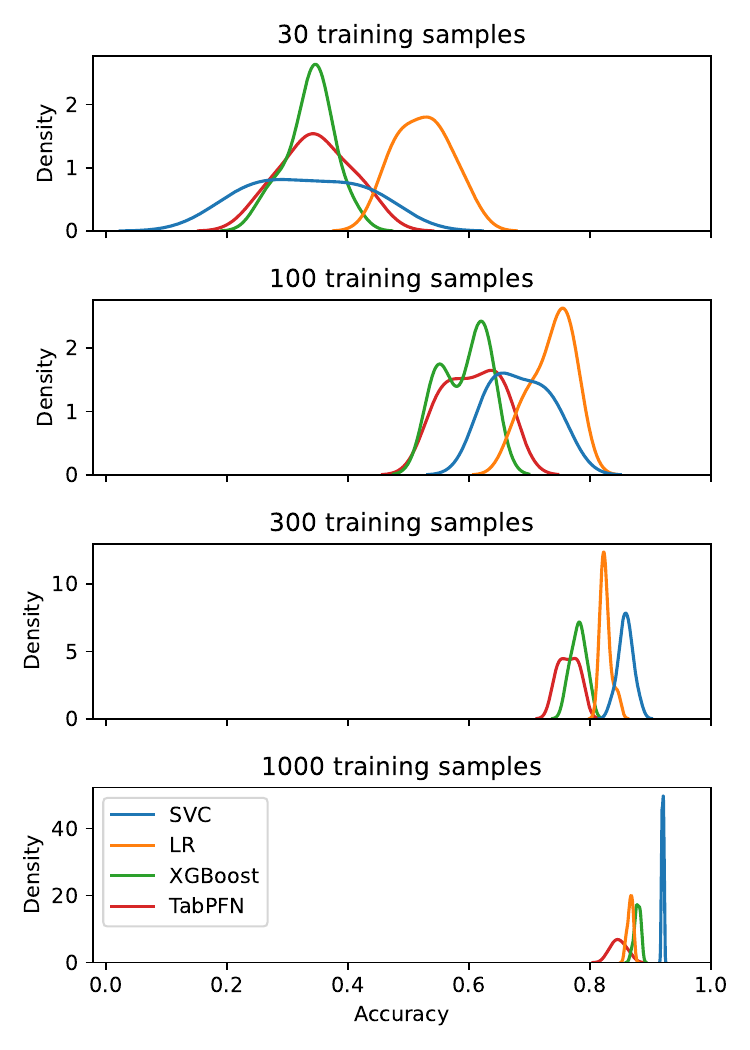} & \includegraphics[scale=0.5]{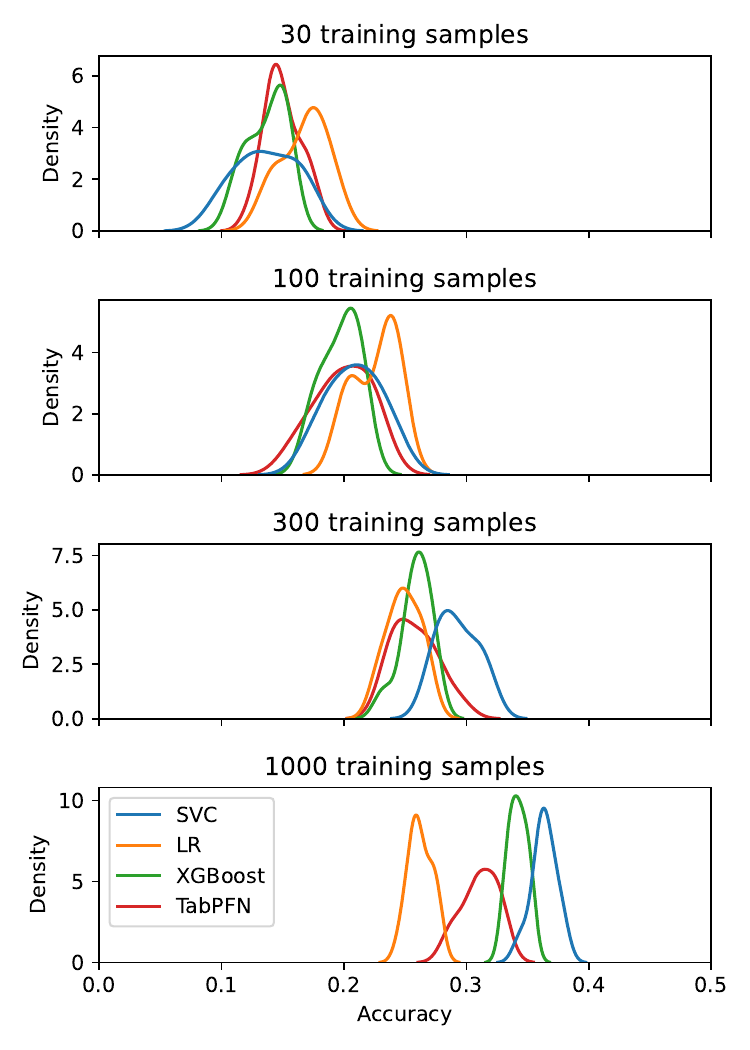} 
    \end{tabular}     
    \caption{Test accuracy on MNIST (left) and CIFAR-10 (right).}
    \label{fig:cv}
\end{figure}

\section{Closing Thoughts}

Taken together, our preliminary results are suggestive of future developments in tabular PFNs. Currently, an applied ML practitioner will likely choose between training a model on their own small dataset and using the TabPFN "one size fits all" model. Our results suggest that TabPFN model will likely perform quite well, even outside its intended domain. However, it still came second-place to logistic regression on our cancer classification task and last or second-last on the CV classification problems. This suggests that the future will not look like a binary choice between training a non-PFN and selecting a single state-of-the-art tabular PFN. Rather, we suspect that there will exist PFNs for specific modalities of data (e.g. gene expression), or for specific settings (e.g. robust classification) that bridge the gap between the two extremes.

In such a future, we believe our approach to evaluating TabPFN will become increasingly essential. In the burgeoning field of large language models (LLMs), evaluation on various public benchmarks is widely considered necessary but insufficient. LLM researchers and users will also evaluate a newly-announced model by trying their favorite personal examples on the new LLM. When the LLM fails on a prompt, one modifies the prompt slightly to see whether the LLM simply expected a different prompting style. When the LLM succeeds, one tries variants to see whether its satisfactory response was in fact brittle to the prompt. By interacting with an LLM, one gets a sense for its expected prompting style and the type of outputs it generates. In particular, providing out-of-distribution (adversarial) inputs (e.g. "poem poem poem") to an LLM tells us something useful about how it will operate on future unanticipated out-of-distribution inputs. 

By analogy, we argue that, while open tabular benchmarks are valuable resources, these should not be fully determinative for researchers and users of tabular ML methods. Benchmarks do allow us to quickly discover which methods are Pareto-dominated and can therefore be safely ignored. However, as we move into a world with multiple available PFN options, with different sorts of inductive priors, it will become increasingly useful to interact with them on simple problems to gain an intuition for whether their priors match one's own use-case. For our analysis on 1d inputs, it is important to notice that there is not necessarily one "right answer". Thus, evaluations of tabular ML approaches will need to be more granular than to describe TabPFN as state of the art for all of tabular ML. Instead, evaluations should aim at identifying specific tabular PFN checkpoints, based on different inductive priors and synthetic datasets, as being best suited for specific classes of problem settings.

Furthermore, our results illuminate a key practical difference between TabPFN, which relies on in-context learning, and other neural network models for tabular ML. Skepticism around neural networks for tabular ML has been justified by problems stemming from the non-convexity of neural network training. Note that the problem (in the small dataset context) with neural net training non-convexity is not fundamentally about the fact that one may have missed a global optimum with better performance. Rather, deep learning requires babysitting during training runs and optimization of training hyperparameters which are unrelated to one's beliefs about the nature of one's specific problem. Thus, a modified architecture, preprocessing method, or data selection approach might be better matched for a particular dataset, but in the end perform worse due to problematic training dynamics -- which one might be unable to fix without risk of overfitting. In the small dataset regime, the maximum performance (over all training hyperparameter settings) matters less than the performance on the default hyperparameter settings. 

Because the overall approach of TabPFN obviates this problem with pure in-context learning, the fundamental weaknesses of other neural network approaches do not apply. For example, our 1d experiments would not have been straightforwardly possible if we had retrained a neural network on each reconfiguration of the training data. If we had done so while keeping the training hyperparameters fixed, it would not represent how people actually use such a neural network. On the other hand, if we had plotted results for carefully optimized hyperparameters, it is not clear whether the results would be illustrative of the general inductive biases of the neural network architecture, or merely of the behavior of an optimally-trained neural network. However, the flip side of this advantage of TabPFN is that our analysis applies not so much to TabPFN-the-method, as it does to \url{https://github.com/automl/TabPFN/blob/d76f4ac7/tabpfn/models_diff/prior_diff_real_checkpoint_n_0_epoch_42.cpkt}-the-checkpoint.

Finally, we believe our evaluation helps address some of the popular skepticism around TabPFN. While our results indicate that there remains substantial room for improvement, we found no evidence that would suggest that TabPFN's results were solely the result of overfitting a large neural network to public benchmarks. Rather, our results suggest that TabPFN learns a simple ``world model'' of small-n statistical learning for tabular classification. This, in itself, makes TabPFN worthy of further careful empirical study.

\bibliographystyle{plainnat}
\bibliography{main}

\appendix

\newpage
\section{Analysis of TabPFN-v2}

First, we show results from running the above analysis on TabPFN-v2.
Then, we show new results which suggest that TabPFN-v2 seems to approximately learn the parity function.

\subsection{Re-analysis of previous experimental settings}

Here we show results from re-running the previous analysis on TabPFN-v2 \citep{hollmann2025accurate}.
The results for the 1d binary classification and 2d multiclass classification experiments show surprisingly worse behavior from TabPFN-v2. 
TabPFN-v2 by default includes hash-based fingerprint features, which allow the model to distinguish duplicated samples.
However, we observe that, with this default setup, the predicted probability functions are now ``bumpy''.

Then, for the real-world non-tabular experiment settings, we were unable to run TabPFN-v2 in a reasonable period of time without causing out-of-memory errors, except on MNIST.
Whereas TabPFN-v1 only used sample embeddings, TabPFN-v2 uses both sample and feature embeddings.
As a result, TabPFN-v2 is more memory-intensive on high-dimensional data.
On MNIST, which we were able to run, TabPFN-v2 improves, coming in the middle-of-the-pack with 300 and 1000 samples, compared to TabPFN-v1 which came in last place on these sample sizes.

\begin{figure}[H]
    \centering
    \includegraphics[scale=0.8]{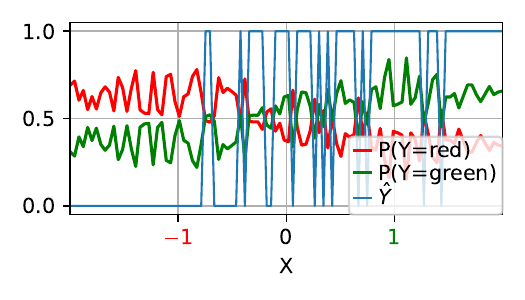}
    \vspace{-0.2in}
    \caption{TabPFN-v2 predicted probabilities for simple 1d scenario, with data in red and green. Compare with Figure \ref{fig:plusminus1-nonmonotone}. }
    \label{fig:v2-plusminus1-nonmonotone}
\end{figure}

\begin{figure}[H]
    \centering
    \includegraphics[scale=0.8]{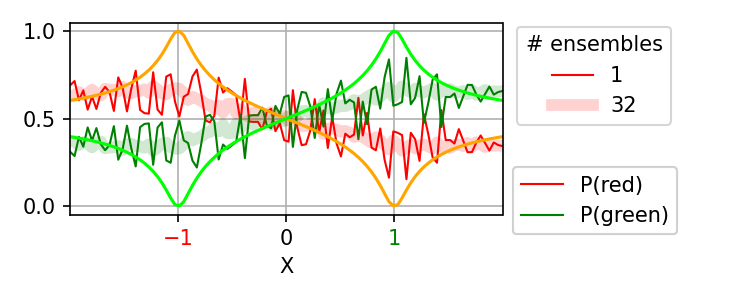}
    \vspace{-0.2in}
    \caption{TabPFN-v2 predicted probabilities for simple 1d scenario, for varying number of ensembles. Also shown are the predicted probabilities from using inverse-square-root of Euclidean distance within softmax, in orange and lime-green. Compare with Figure {fig:plusminus1-ensembles2}.}
    \label{fig:v2-plusminus1-ensembles2}
\end{figure}

\begin{figure}[H]
    \centering
    \includegraphics[scale=0.7]{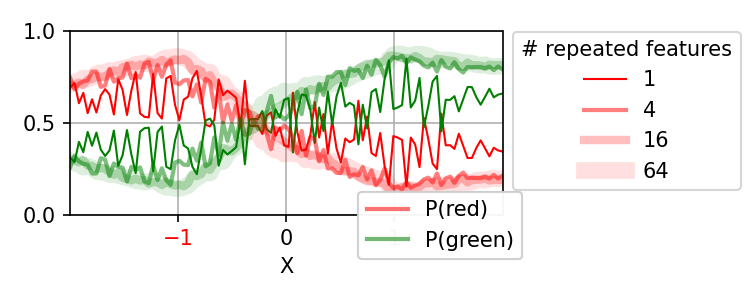}
    \vspace{-0.2in}
    \caption{TabPFN-v2 predicted probabilities for simple 1d scenario, but with repeated features. Compare with Figure \ref{fig:plusminus1-repeat-features}.}
    \label{fig:v2-plusminus1-repeat-features}
\end{figure}

\begin{figure}[H]
    \centering
    \includegraphics[scale=0.7]{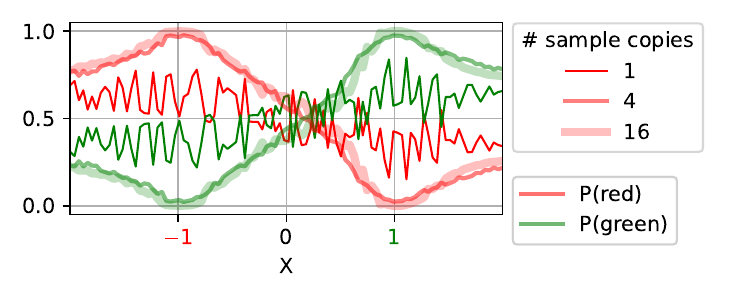}
    \vspace{-0.2in}
    \caption{TabPFN-v2 predicted probabilities for simple 1d scenario, but when both red and green samples are duplicated. Compare with Figure \ref{fig:plusminus1-repeat-samples}.}
    \label{fig:v2-plusminus1-repeat-samples}
\end{figure}

\begin{figure}[H]
    \centering
    \includegraphics[scale=0.7]{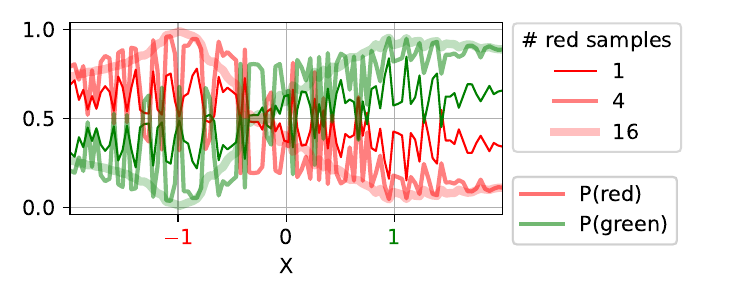}
    \vspace{-0.2in}
    \caption{TabPFN-v2 predicted probabilities for simple 1d scenario, but when the red sample is duplicated. Compare with Figure \ref{fig:plusminus1-repeat-red}.}
    \label{fig:v2-plusminus1-repeat-red}
\end{figure}

\begin{figure}[H]
    \centering
    \includegraphics[scale=0.7]{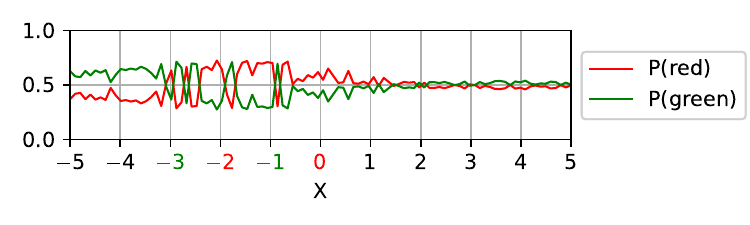} \\
    \includegraphics[scale=0.7]{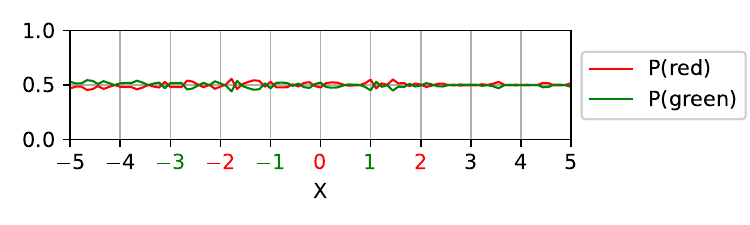} \\
    \includegraphics[scale=0.7]{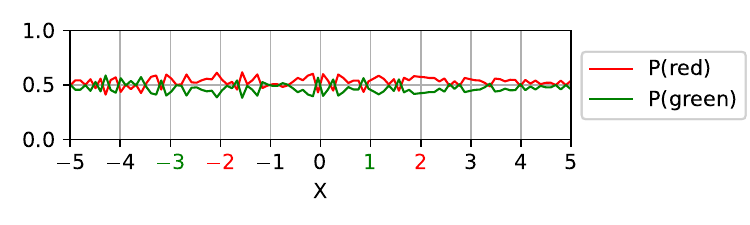}
    \vspace{-0.2in}
    \caption{TabPFN-v2 predicted probabilities for three scenarios with periodic patterns. }
    \label{fig:v2-plusminus1-periodic}
\end{figure}

\begin{figure}[H]
    \centering
    \includegraphics[scale=0.5]{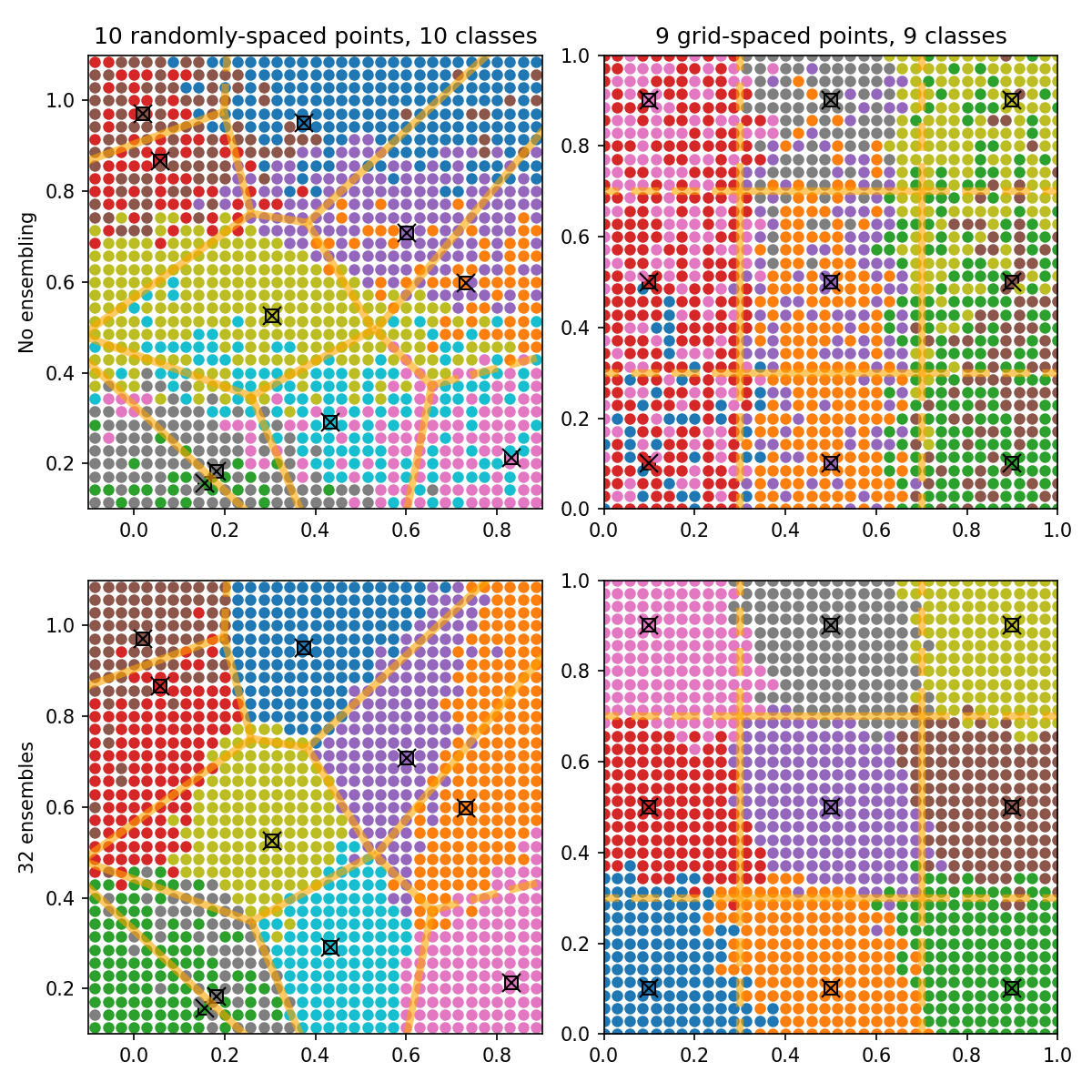}
    \vspace{-0.2in}
    \caption{TabPFN-v2 predictions on randomly-spaced points (left) and grid-spaced points (right). The training points are depicted as $\times$s. The yellow lines depict the Voronoi diagram of the training points. The test points are colored by TabPFN's predictions, using the same color scheme as the training points. Compare to Figure \ref{fig:voronoi} for TabPFN-v1.}
    \label{fig:v2-voronoi}
\end{figure}
\begin{figure}[H]
    \centering   
    \begin{tabular}{cc}
        %MNIST &  \\
         \includegraphics[scale=0.5]{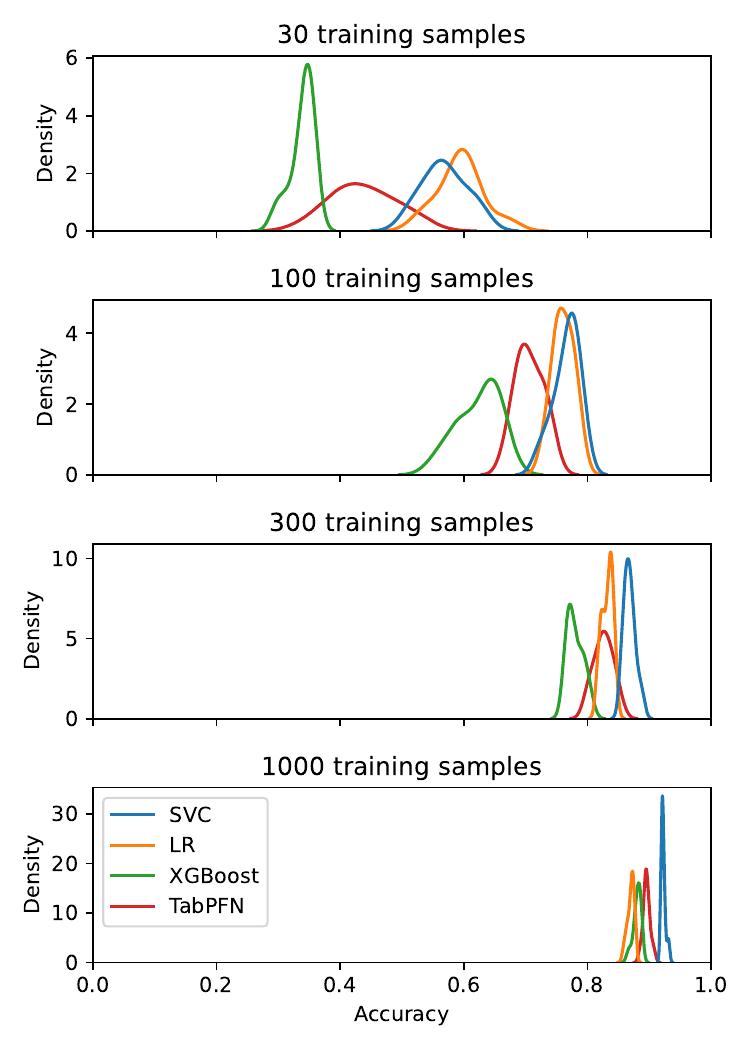} &  
    \end{tabular}     
    \caption{Test accuracy on MNIST. Compare to Figure \ref{fig:cv}.}
    \label{fig:v2-cv}
\end{figure}

\subsection{TabPFN-v2 approximately learns the parity function}

Here we report our findings from training TabPFN-v2 on truth tables for the parity function. We evaluate four methods, XGBoost \citep{chen2016xgboost}, and TabPFN-v2 with 1, 4, and 16 estimators. We vary the number of input dimensions $D$ from 3 to 10, which yields truth tables with $2^3$ to $2^{10}$ examples, respectively.
For each truth table of dimension $D$, we exhaustively evaluate the performance on all possible $2^D$ folds, comprising single-example test sets.
We then train each of the four classifiers above to predict the parity function computed on these inputs, and report the performance averaged over all $2^D$ test sets).
We consider four training regimes: using all non-test examples (equivalent to leave-one-out evaluation) for training, using 50\% of non-test examples (i.e. $\lfloor \frac{2^D-1}{2} \rfloor$ per fold), using 25\% of non-test examples, and using a constant 127 training examples.

Results are shown in Figure \ref{fig:parity}. We see that TabPFN-v2 with ensembling seems to be able to approximately learn the parity function, with improving accuracy as the number of dimensions increases.
(Note that the exponential convergence in dimensions corresponds to linear convergence in number of examples.)
Impressively, TabPFN-v2 seems to have the same convergence in error rate when only 50\% of samples are provided as training data.
Furthermore, with a constant number (127) of training examples, TabPFN-v2 manages to improve as the number of dimensions increases from 7 to 9, before starting to worsen at 10 dimensions.
Still, TabPFN-v2's performance with 127 training examples at dimensionality of 10 is quite impressive: with roughly 12\% of the rows of the truth table, TabPFN-v2 predicts the parity function with >99\% accuracy!

\iffalse
\begin{figure}[H]
    %\begin{tabular}{cc}
    (A) Training on all non-test examples \\
     \includegraphics[scale=0.8]{v2figures/parity-error-Nminus1.pdf}\\
     (B) Training on 50\% of non-test examples \\
     \includegraphics[scale=0.8]{v2figures/parity-error-half.pdf}\\     
     (C) Training on 25\% of non-test examples \\
     \includegraphics[scale=0.8]{v2figures/parity-error-quarter.pdf}\\
     (D) Training on 127 non-test examples \\
    \includegraphics[scale=0.8]{v2figures/parity-error-constant.pdf}\\     
    %\end{tabular}     
    \caption{Parity-learning results. On the y-axes are shown, in log scale, the test-set error rate, which is 1 minus the (mean) accuracy. Results are shown for (A) training on all non-test examples, (B) training on 50\% of non-test examples, (C) training on 25\% of non-test examples, and (D) training on 127 non-test examples. Note that for training on 127 non-test examples in (D), we are only able to evaluate methods on truth tables with at least $2^7$ rows.}
    \label{fig:parity}
\end{figure}
\fi

\begin{figure}[H]
    \begin{tabular}{ll}
    (A) & (B)  \\ 
    \includegraphics[scale=0.75,clip=true,trim=0 0 2.25in 0]{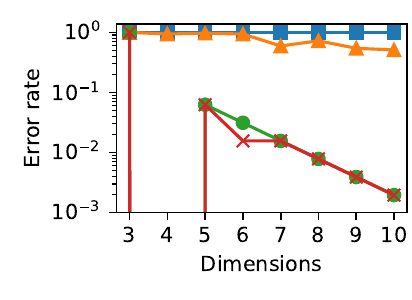} &
     \includegraphics[scale=0.75]{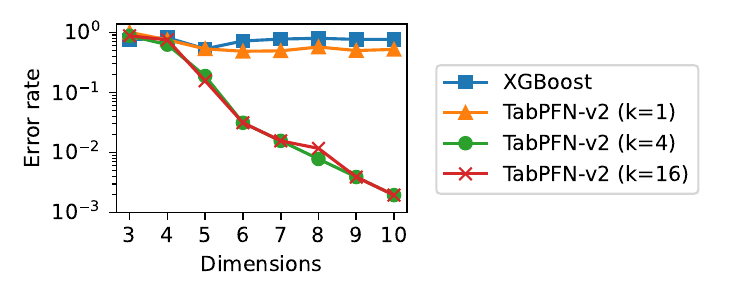} \\
     (C) & (D) \\
     \includegraphics[scale=0.75,clip=true,trim=0 0 2.25in 0]{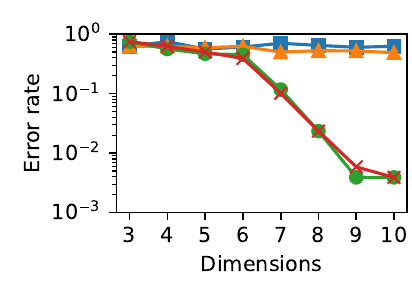} &
    \includegraphics[scale=0.75]{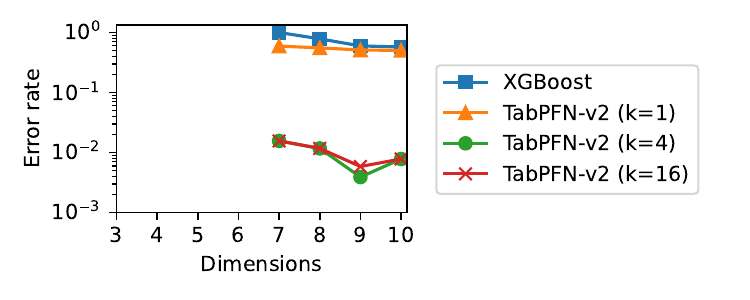}\\     
    \end{tabular}     
    \caption{Parity-learning results. On the y-axes are shown, in log scale, the test-set error rate, which is 1 minus the (mean) accuracy. Results are shown for (A) training on all non-test examples, (B) training on 50\% of non-test examples, (C) training on 25\% of non-test examples, and (D) training on 127 non-test examples. Note that for training on 127 non-test examples in (D), we are only able to evaluate methods on truth tables with at least $2^7$ rows.}
    \label{fig:parity}
\end{figure}

\end{document}